\documentclass[pdflatex,sn-basic]{sn-jnl}

\usepackage{hyperref}
\usepackage{amsmath,amssymb,amsfonts,amsthm}
\usepackage{mathtools}
\usepackage{graphicx}
\usepackage{textcomp}
\usepackage{xcolor}
\usepackage{tabularx} 
\usepackage{bbm}

\usepackage{algorithm, algorithmicx}
\usepackage[switch]{lineno}
\usepackage{amsfonts}
\usepackage{adjustbox}
\usepackage{enumitem}
\theoremstyle{plain}

\newtheorem{theorem}{Theorem}[section]

\def\BibTeX{{\rm B\kern-.05em{\sc i\kern-.025em b}\kern-.08em
    T\kern-.1667em\lower.7ex\hbox{E}\kern-.125emX}}
    
\newcolumntype{C}{>{\centering\arraybackslash}X}
\newcolumntype{P}[1]{>{\centering\arraybackslash}p{#1}} 

\jyear{2022}%

\begin{document}

\title[From Primes to Paths: Enabling Fast Multi-Relational Graph Analysis]{From Primes to Paths: Enabling Fast Multi-Relational Graph Analysis}



\author*[1,2]{\fnm{Konstantinos } \sur{Bougiatiotis}}\email{ \small kbogas@\{di.uoa.gr;iit.demokritos.gr\}}

\author[1]{\fnm{Georgios} \sur{Paliouras}}

\affil*[1]{\orgdiv{Institute of Informatics and Telecommunications}, \orgname{National Center Scientific Research Demokritos}, \orgaddress{\city{Athens} \country{Greece}}}

\affil[2]{\orgdiv{Department of Informatics and Telecommunications}, \orgname{National and Kapodistrian University}, \orgaddress{\city{Athens} \country{Greece}}}

\abstract{Multi-relational networks capture intricate relationships in data and have diverse applications across fields such as biomedical, financial, and social sciences. As networks derived from increasingly large datasets become more common, identifying efficient methods for representing and analyzing them becomes crucial. This work extends the Prime Adjacency Matrices (PAMs) framework, which employs prime numbers to represent distinct relations within a network uniquely. This enables a compact representation of a complete multi-relational graph using a single adjacency matrix, which, in turn, facilitates quick computation of multi-hop adjacency matrices. In this work, we enhance the framework by introducing a lossless algorithm for calculating the multi-hop matrices and propose the Bag of Paths (BoP) representation, a versatile feature extraction methodology for various graph analytics tasks, at the node, edge, and graph level. We demonstrate the efficiency of the framework across various tasks and datasets, showing that simple BoP-based models perform comparably to or better than commonly used neural models while offering improved speed and interpretability.}




\maketitle

\section{Introduction}

In recent years, research on complex networks has significantly matured, drawing attention across various domains such as biological, social, and financial networks, among others~\citep{boccaletti2006complex}. This is primarily due to their ability to model intricate relationships within data, making them invaluable in real-world scenarios, where complex structures are prevalent. A significant subdomain in this field is the study of multi-relational networks, which consider that entities (i.e. nodes) in a complex network can be connected through multiple types of links. Multi-relational networks come in different  variants, such as multi-layer, multi-dimensional or multi-plex networks, and knowledge graphs~\citep{zou2020survey}. A good overview of these naming conventions, their definitions, and differences can be found in ~\citep{kivela2014multilayer}. In this work, we use the term multi-relational graph/network as an umbrella term, encompassing all complex networks that can be represented as a collection of triples in the form $(s, r, o)$. In this form, $s$ and $o$ represent subject and object entities, while $r$ denotes the relation connecting them.

One of the main objectives of research in this area is to generate insights, by aggregating information expressed through each relation, without losing essential details. Various approaches address this problem for different downstream tasks, including embedding techniques for nodes and relations,~\citep{wang2017knowledge}, tensor decompositions~\citep{kolda2009tensor}, symbolic methodologies~\citep{ji2021survey} and more recently graph neural networks~\citep{zhou2020graph}. However, many of these approaches focus solely on direct relations between entities, often neglecting the richer context provided by multi-hop connections within the graph~\citep{sato2020survey}. In numerous domains, such as explainable AI and molecular chemistry~\citep{edwards2021explainable, liu2014assessment}, understanding the paths connecting entities can reveal the true nature of their relationships, the roles of each entity in the graph, and ultimately aid in the task at hand. Therefore, there is a compelling need for a framework that facilitates the efficient computation of representations capturing the multi-hop information within a network.

To address this need, the \textit{Prime Adjacency Matrix} (PAM) framework was initially presented in \citep{bougiatiotis_cbms}. This innovative representation compacts all relations of the original network into a single adjacency matrix in a lossless manner. By leveraging the unique factorization properties of integers, each relation type is mapped to a distinct prime number, allowing the construction of an adjacency matrix, that encapsulates all the information of the original graph. This consolidated adjacency matrix enables the swift computation of multi-hop adjacency matrices, essential for extracting rich relational data across multiple hops in the graph. The process is computationally efficient and scalable, making it applicable to large, complex networks common in various real-world applications.

Building on the original proof-of-concept work, we significantly extend the PAM framework to improve its performance and utility. Firstly, we introduce an algorithm for the calculation of higher-order matrices that ensures multi-hop paths can be extracted without loss of information. Secondly, we propose the Bag of Paths (BoP) feature extraction methodology, which generates versatile and interpretable feature vectors for node, edge, and graph-level analytics. This rich structural information is utilized across a wide range of tasks. Finally, we implement the framework using GraphBLAS~\citep{erik_welch_2024_10631255}, significantly improving computational efficiency.

To conclude, the main contributions of this work are the following:
\begin{itemize}
    \item Development of the lossless algorithm for generating $k$-hop matrices, alongside experiments for validation.
    \item Introducing the Bag of Paths representation, a common feature extraction methodology to address graph-related tasks at multiple scales, based on efficient matrix look-ups that allow for interpretable path-based features.
    \item Extended the application of the PAM framework on node and graph level tasks, demonstrating its effectiveness in terms of speed and competitive performance across multiple scenarios. 
    \item Optimized the implementation of the framework utilizing GraphBLAS, greatly improving performance and making it accessible as a python module\footnote{\url{https://pypi.org/project/prime-adj/}} for further experimentation.
\end{itemize}

The rest of the paper is structured as follows: 
Section~\ref{sec:methodology} introduces the framework in detail. Then we present its application on different tasks in Section~\ref{sec:applications}, with experimental results and insights into the framework's capabilities. Section~\ref{sec:related_work} refers to related work on path-based approaches and a discussion on the usability of PAMs. Finally, we summarize the main aspects of the framework and possible future directions in Section~\ref{sec:conclusions}.\footnote{The code and related scripts are available at \url{https://github.com/kbogas/PAM_BoP}.}

\section{Methodology}\label{sec:methodology}
In this section, we provide an overview of the Prime Adjacency Matrices framework, emphasizing its key features. Parts of this introduction were presented in a different form in~\citep{bougiatiotis_cn}. Then, we extend the framework with a lossless $k$-hop variant, providing a high-level description of the algorithm in Section~\ref{subsec:lossless_k_hop}, while a detailed proof is provided in Appendix~\ref{sec:appendix_lossless}. Moreover, we present a simple, fast, and generic feature extraction methodology based on PAMs in Section~\ref{subsec:BoP}, which is used for multiple downstream tasks.

\subsection{Definition}
We start with an unweighted, directed, multi-relational graph $\mathcal{G}$, with $V$ nodes, $E$ edges, and $R$ unique relation types, denoted as $\mathcal{G}=(V,E,R)$ with $E \subseteq V\times V\times R$ and $\lvert V \rvert = N$. Typically, the graph can be represented as an adjacency tensor $A$ of shape $N \times N \times R$:
\begin{equation}\label{eq:tensor_A}
        A[i,j,r]=
        \begin{cases}
            1 & \text{if $r$ connects nodes $i,j$}\\
            0 &\text{otherwise}
        \end{cases}
\end{equation}

We map each unique relation type $r \in R$ to a distinct prime number $p_r$ through a mapping function $\varphi$, such that: $ \forall r \in R: \varphi(r) = p_r$, where $p_r$ is prime and $p_i = p_j \iff i = j$. This mapping function is a design choice and at its simplest form, we can randomly order the relations and allocate each relation to the next available prime, starting from the number $2$.

With this mapping, we construct the \textit{Prime Adjacency Matrix} (PAM) $P$ of shape $N \times N$ as follows:

\begin{equation}\label{eq:PAM}
        P[i,j]=
        \begin{cases}
             \displaystyle \prod_{r:A[i,j,r]=1} p_r & if\text{ $\exists r : A[i,j,r]=1 $} \vspace{0.3em}\\ 
             \hspace{2em} $0$ &if\text{ $\forall r :$ $A[i,j,r]=0$}
        \end{cases}
\end{equation}

In Eq.~\eqref{eq:PAM}, each non-zero element $P[i,j]$ is the product of the primes $p_r$ for all relations $r$ that connect node $i$ to node $j$. Using the Fundamental Theorem of Arithmetic (FTA)~(Appendix~\ref{theorem:FTA}), we can decompose each product to its prime factors, preserving the full structure of $\mathcal{G}$ in $P$ without any loss.

We also define $P_{+}$, a variant that sums the relations between two nodes instead of multiplying them, as follows:
\begin{equation}\label{eq:PAM_plus}
        P_{+}[i,j]=
        \begin{cases}
             \displaystyle \sum_{r:A[i,j,r]=1} p_r & if\text{ $\exists r : A[i,j,r]=1 $}\vspace{0.3em}\\
             \hspace{2em} $0$ &if\text{ $\forall r :$ $A[i,j,r]=0$}
        \end{cases}
\end{equation}

This variant is not lossless, but it is convenient for use in applications. If each pair of nodes $i, j$ in $\mathcal{G}$ has at most one relation between them, $P$ and $P_{+}$ are identical.

\subsection{The one-hop case}

\begin{figure}[htbp]

\centerline{\includegraphics{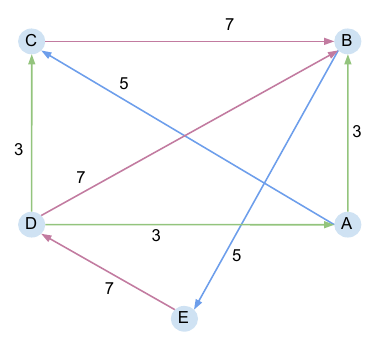}}

\caption{An example multi-relational graph with 5 nodes and 3 types of relation.}
\label{fig:PAM_rolling_example}
\end{figure}

Consider a case where each pair of nodes is connected by at most one relation. In such cases, the values in $P[i,j]$ are simply the corresponding $p_r$ numbers that connect $(i,j)$. A small graph of this type, with five nodes, is shown in Fig.~\ref{fig:PAM_rolling_example}. It has three types of relations mapped to $3$ ($r_1$, green), $5$ ($r_2$, blue), and $7$ ($r_3$, magenta).

The resulting PAM would be:

\vspace{0.06cm}$P=
\left\{\begin{smallmatrix}
0 & 3 & 5 & 0 & 0 \\
0 & 0 & 0 & 0 & 5 \\
0 & 7 & 0 & 0 & 0 \\
3 & 7 & 3 & 0 & 0 \\
0 & 0 & 0 & 7 & 0 \\
\end{smallmatrix}\right\},$\vspace{0.06cm} 

\noindent with node A corresponding to index 0, node B to index 1, and so forth. For the edge $A \xrightarrow{3} B$, $P[0,1] = 3$; for $A \xrightarrow{5} C$, $P[0,2] = 5$, and so on, expressing all edges in the graph.

This compact PAM representation allows interesting observations regarding the structure and the types of edges in the graph. For example, by examining $P[0,:]$ and $P[:,0]$, we see that node A has two outgoing edges (types $3$ and $5$) and one incoming edge (type $3$). Another easily inferred graph property is the frequency of different relations, obtained by counting the occurrences of the non-zero elements of $P$, yielding the distribution $\{\mathit{3}:3, \mathit{5}:2, \mathit{7}:3\}$, where $\{\mathit{3}:3\}$ denotes that relation $\mathit{3}$ is exhibited $3$ times, and so forth.


\subsection{The multi-hop case}

Having a single adjacency matrix for the entire graph $\mathcal{G}$ allows us to use tools from classical network analysis. For instance, we can easily obtain the powers of the adjacency matrix. In a single-relational, directed and unweighted network, the element $(i, j)$ of the power $k$ of an adjacency matrix represents the number of paths of length $k$ from node $i$ to node $j$. Generalizing this property in PAMs, where each value also represents a specific relation type, the values of $P^k[i,j]$ track the relational chains linking two nodes. To see this, consider the second-order PAM for the  graph of Fig.~\ref{fig:PAM_rolling_example}:\newline
$P^{2}= P\times P =
\left\{ \begin{smallmatrix}
0 & 35 & 0 & 0 & 15 \\
0 & 0 & 0 & 35 & 0 \\
0 & 0 & 0 & 0 & 35 \\
0 & 30 & 15 & 0 & 35 \\
21 & 49 & 21 & 0 & 0 \\
\end{smallmatrix}\right\}.$\vspace{0.06cm}

For the node pair $(A, B)$, we have $P^2[0,1] = 35$. From Fig.~\ref{fig:PAM_rolling_example}, we see that $A$ connects to $B$ in two hops via node $C$, following the path $A \xrightarrow{5} C \xrightarrow{7} B$. The relations $5$ and $7$ in this path are encapsulated in $P^2[A,B] = 35 = 5\times7$, as the prime factors of the cell value. Similarly, $P^2[A,E] = 15 = 3 \times 5$ corresponds to $A \xrightarrow{3} B \xrightarrow{5} E$, $P^2[E,A] = 21 = 7 \times 3$ corresponds to $E \xrightarrow{7} B \xrightarrow{3} A$, and so forth. Therefore, the products in $P^k$ express the relational $k$-chains linking nodes in the graph.

Importantly, $P^2[D, B] = 30$ is the sum of two possible paths $30 = 9 + 21 = 3\times3 + 3\times7$, corresponding to $D \xrightarrow{3} A \xrightarrow{3} B$ and $D \xrightarrow{3} C \xrightarrow{7} B$. This shows that each cell $(i, j)$ in $P^k$ aggregates all ``path-products'' of $k$-hops from $i$ to $j$. This is in line with the concept of adjacency matrix powers in classical graph theory, while also encoding relation types in the cell values.


We can also extract structural characteristics for elements of the graph (e.g. nodes, subgraphs etc.) by examining this $P^2$. For instance, the frequency of $2$-hop paths is determined by counting the non-zero values in $P^2$, yielding $\{\mathit{15}:2, \mathit{21}:2, \mathit{30}:1, \mathit{35}:4, \mathit{49}:1\}$. These frequencies can then be used in further analysis. For example, in a molecular graph, where atoms are nodes and bonds are relational edges, a frequent pattern like $\mathit{35}$, corresponding to specific bonds mapped to the primes $\{5, 7\}$, could be important for characterizing the molecule’s properties such as toxicity or solubility~\citep{sharma2017toxim}. We can extract from the 2-hop PAM, similar information as we extract from the 1-hop PAM, such as the incoming/outgoing 2-hop paths for each node.

This procedure can be repeated for as many hops as needed by calculating the corresponding $P^k$. The values in these matrices contain aggregated information regarding the relational chains of length $k$ that connect the corresponding nodes, allowing easy extraction of interesting characteristics through simple operations.

\subsection{The multigraph case}

\begin{figure}[htbp]
\centerline{\includegraphics{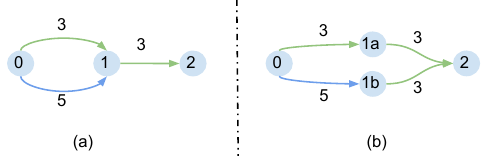}}
\caption{(a) Simple graph with multiple relations between two nodes. (b) The same graph with node 1 split into 2 substitute nodes, namely 1a and 1b.}
\label{fig:PAM_2hop_example}
\end{figure}

Now, let us consider the more general case where multiple relations exist between a pair of nodes. These complex networks are of interest to many domains~\citep{battiston2017multilayer, gallotti2016lost, boccaletti2014structure}. The small graph shown in Fig.~\ref{fig:PAM_2hop_example}~(a) has 3 nodes and 2 different types of relations, green and blue, mapped to the numbers $3$ and $5$, respectively. Moreover, nodes $(0,1)$ are connected with both relations $3$ and $5$ simultaneously.

Let us focus on the node pair $(0,2)$. The corresponding values of the 2-hop adjacency matrix should express the sum of the 2-hop paths between these two nodes. We expect the 2-hop value of $P^2$ to be $P^2[0,2]= 3\times3 + 5\times3 = 24$, aggregating the paths $0\xrightarrow{3}1\xrightarrow{3}2$ and $0\xrightarrow{5}1\xrightarrow{3}2$. To visualize these two paths, we could split node 1 into two distinct nodes (1a) and (1b) as shown in Fig.~\ref{fig:PAM_2hop_example}~(b).

However, the PAM for this graph according to Eq.~\eqref{eq:PAM} would be:\vspace{0.05cm} $P = \left\{ \begin{smallmatrix}
0 & 15 & 0\\
0 & 0 & 3 \\
0 & 0 & 0 \\
\end{smallmatrix}\right\} $ and the corresponding $P^2 =  P \times P = \left\{\begin{smallmatrix}
0 & 0 & 45\\
0 & 0 & 0 \\
0 & 0 & 0 \\
\end{smallmatrix}\right\}$.\vspace{0.05cm} We can see that $P[0,1] = 3\times5 = 15$,  representing the product of the relations between the nodes. This leads to $P^2[0,2] = 45 = 15\times3 = P[0,1]\times P[1,2]$, which is different from the expected value of $24$.

The expected value can be achieved if summation is used to aggregate paths in $P$ instead of  multiplication (i.e. using $P_{+}$ instead of $P$). Then the corresponding PAM, with the sum of the primes instead of their product, would be \vspace{0.1cm}$P_{+}=
\left\{ \begin{smallmatrix}
0 & 8 & 0\\
0 & 0 & 3 \\
0 & 0 & 0 \\
\end{smallmatrix}\right\}$ and the corresponding $P_{+}^2$ would be $P_{+}^2 = P_{+} \times P_{+} = \left\{ \begin{smallmatrix}
0 & 0 & 24\\
0 & 0 & 0 \\
0 & 0 & 0 \\
\end{smallmatrix}\right\}$\vspace{0.05cm}.


These two different aggregation approaches reveal two different concepts, namely ``extension of an existing path'' and the ``aggregation of a collection of paths''. In standard matrix multiplication, the multiplication operation is used for path extension. For example, going back to the example of Fig.~\ref{fig:PAM_rolling_example}, we had $P^2[A,B]= 35 = 5 \times 7$, where we extend the path  $A \xrightarrow{5} C$ with $C \xrightarrow{7} B$, by multiplying the primes of the relations, creating a value that represents the chain of relations. On the other hand, for the aggregation of paths, we use the summation of their resulting values, as pointed out in the case of $P^2[D, B] = 30 = 9 + 21 = 3\times3 + 3\times7$, where two path values have been merged in a single value. 

Therefore, to conform with the notion that the aggregation of paths is done through summation, we will be using $P_{+}$ as introduced in Eq.~\eqref{eq:PAM_plus}, where the sum of the primes is used, instead of their product. This representation is not lossless, as the sum of the primes cannot be uniquely decomposed back to the original primes, unlike their product. So, if we need to represent the full graph $\mathcal{G}$ without loss using an adjacency matrix, we need to  use the original $P$ from Eq.~\eqref{eq:PAM}. 
In  the rest of the paper, when a power of PAM is used and presented as $P^{k}$, it will be calculated using $P_{+}$, unless mentioned otherwise.

\subsection{The k-hop lossless case}\label{subsec:lossless_k_hop}
The distinction between path extension and path aggregation is useful in developing  a process for generating lossless k-hop PAMs. In our setup, lossless means that given a value $P^k[i,j]$, we can extract all the paths of $k$-hops that connect $i$ to $j$ without any loss of information.

To create such a process, we need a way to ``extend existing paths with new relations''
(chaining) and ``aggregate collections of paths'' (aggregating) as before, but without losing information. Through standard matrix multiplication, we introduce uncertainty (loss of information) in the final result. For example, multiplying relations obscures their order due to the commutativity of the operation, e.g., $15 = 3 \times 5 = 5 \times 3$. Also, retrieving the initial summands that aggregate to a final value is not trivial.

We first define a lossless way to aggregate a collection of $k$-hop paths into a single value.
Assume there are $M$ directed, $k$-hop paths connecting nodes $i$ and $j$. We denote them as follows:

\begin{equation}\label{eq:path_def}
Paths^k[i,j] = \underbrace{\{path_1, path_2, ...\}}_{M} = \underbrace{\{\overbrace{(r_1, r_2, ...)}^{k}, \overbrace{(r_1, r_2, ...)}^{k}, ...\}}_{M}
\end{equation}

Using this notation, the lossless aggregation process is shown below.\newline

\noindent\underline{\textit{Aggregation Process (AP)}}\newline
\begin{enumerate}[label=AP Step \arabic*.,itemindent=*]
    \item For a given matrix order $k$, map each possible path $path_i$ of $k$-hops to a corresponding prime $p_i$\footnote{The mapping procedure $\phi_k$ is not crucial for the algorithm. We can start mapping each distinct path $\overbrace{(r_1, r_1, ...)}^{k}, \overbrace{(r_1, r_2, ...)}^{k}$ sequentially to 2, 3, ..., and so on, until we've exhausted all possible $k$-hop relational chains. For $\lvert R \rvert$ distinct relations in the graph, the possible combinations at $k$-hops yield $\lvert R\rvert^k$ distinct relational chains, so we will need at most a mapping with $\lvert R\rvert^k$ entries.}. That is: 
    \begin{equation}\label{eq:paths_mapped_single}
     \forall \texttt{ $path_i \in Paths^k$ , } path_i \xrightarrow{\phi_k} p_i
    \end{equation}
    \item Given a collection of paths as in Eq.~\eqref{eq:path_def}, we can \textbf{uniquely} represent them using a single value, through the Fundamental Theorem of Arithmetic~(Appendix~\ref{theorem:FTA}), as the product of the primes that correspond to the paths in the collection. That is: 
    \begin{equation}\label{eq:paths_mapped_product}
     Paths^k[i,j] = \underbrace{\{path_1, path_2, ...\}}_{M} \overset{\mathrm{AP-1}}{=} \underbrace{\{p_1, p_2 ,...\}}_{M} = \mathop{\prod}_{i=1}^M p_i = P^k[i,j]
    \end{equation}
\end{enumerate}

\noindent \newline 
\indent In this way the final value $ P^k[i,j]$ can be decomposed to its prime factors, which then  can be mapped through $\phi{_k}^{-1}$ back to the original $Paths^k$.  

\noindent With the lossless aggregation process in place, we can now design the extension process. To construct, $P^{k+1}[i,j]$ we need all the $(k+1)$-hop paths that connect $i$ to $j$:

\begin{equation}\label{eq:pij_k1}
    P^{k+1}[i,j] = \{path_1, path_2 ...\} =\{\overbrace{(r_1, r_2, ...)}^{k+1}, \overbrace{(r_1, r_2, ...)}^{k+1}, ...\}
\end{equation}

Each of these $(k+1)$-hop paths can be broken into two parts. The first part denotes the $k$-hop path starting from node $i$ that reaches an intermediate node $n_c$, and the second part is the final hop from $n_c$ to the tail node $j$ (where the intermediate node may be different in each path):

\begin{equation}\label{eq:path_segm}
    path_i = \overbrace{(r_1, r_2, ...)}^{k+1} = \overbrace{(r_1,r_2,...,r_k)}^{\text{first part}} \lvert\rvert \overbrace{(r_{k+1})}^{\text{second part}} 
\end{equation}
where $\lvert\rvert$ denotes the concatenation of the path with the final relation.

We can extract all $(k+1)$-hop paths that connect $i,j$ by finding all $k$-hop paths starting from $i$ and leading up to any $n_c$ in $\mathcal{G}$, under the constraint that there exists also a 1-hop connection between $n_c$ and $j$. That is:

\begin{equation}\label{eq:path_k+1_def}
\small
Paths^{k+1}[i,j] = \{Paths^k[i,n_c]\lvert \rvert Paths^1[n_c,j] : n_c \in V, P^k[i,n_c]*P[n_c, j] \neq 0\}
\end{equation}

Based on the above intuition, we can design the lossless chaining process. For a given $k$-hop value $P^k[i,n_c]$, the final hop $P[n_c, j]$, and the mapping functions $\phi_k, \phi_1$, the process is described below.\newline

\noindent\underline{\textit{Chaining Process (CP)}}\newline
\begin{enumerate}[label=CP Step\arabic*.,itemindent=*]
    \item Get the prime factors of the values $P^k[i,n_c]$ and $P[n_c, j]$
    \item Map the prime factors back to paths using $\phi_k^{-1}$ and $\phi_1^{-1}$ accordingly
    \item Extend each of the $k$-hop paths $(r_1,r_2,...,r_k)$ with each final 1-hop path $r_{k+1}$ by concatenation: $(r_1,r_2,...,r_{k+1})$
\end{enumerate}

\noindent \newline 
\noindent Using the above chaining process, no information regarding the paths is lost, and we can generate $Paths^{k+1}[i,j]$ by extending the paths $Paths^k[i,n_c]$ with $Paths^1[n_c,j]$, for all appropriate intermediate nodes $n_c$.

\begin{figure}[htbp]
\centerline{\includegraphics[width=1\textwidth]{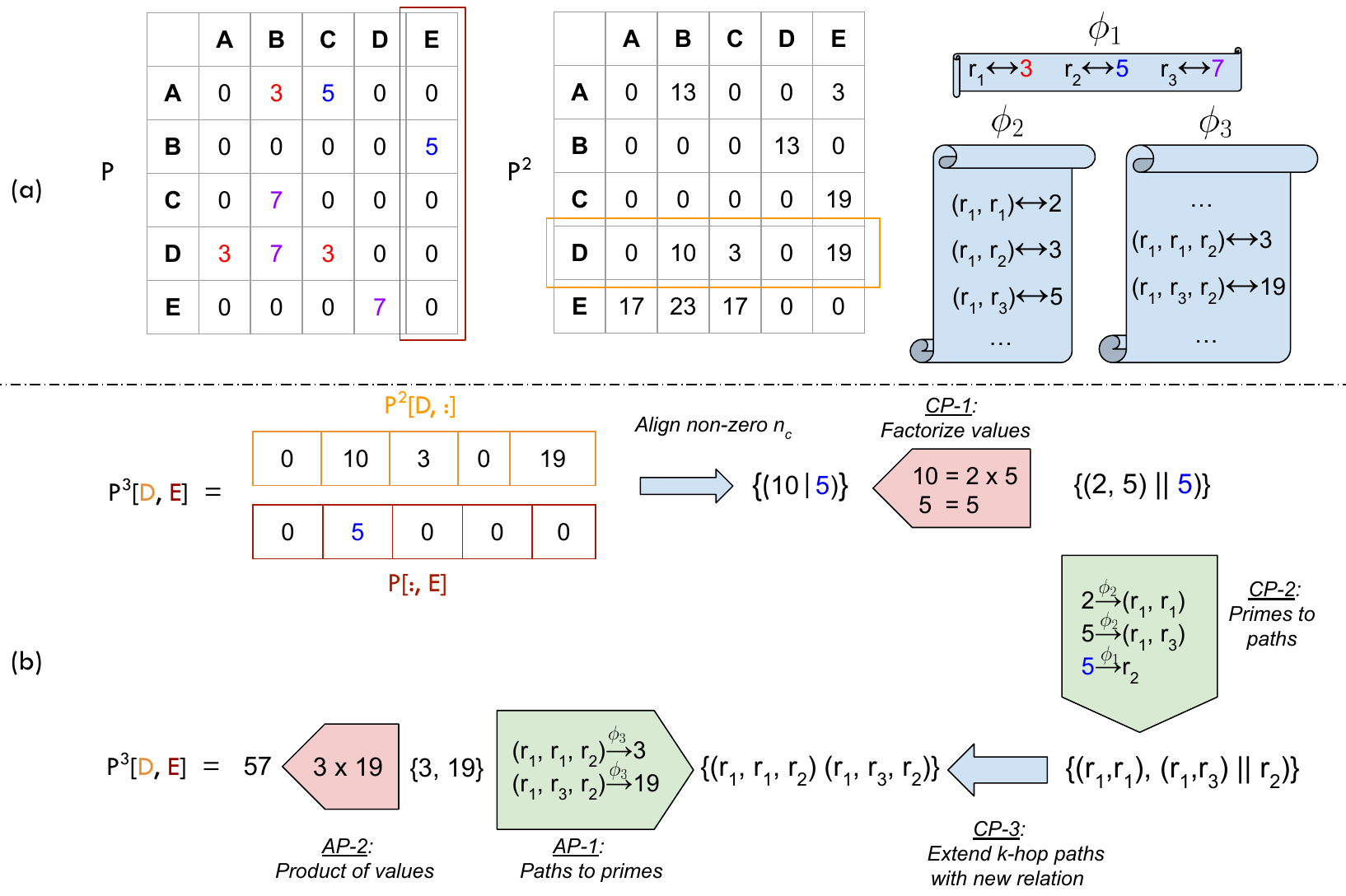}}
\caption{Illustration of the lossless generation of the value for $P^3[D,E]$, corresponding to two $3$-hop paths. (a) The lossless $P, P^2$ PAMs and the mappings $\phi_{1},\phi_{2},\phi_{3}$ that will be used. (b) The process of generating $P^3[D,E]$ without loss, through the aggregation (AP) and chaining (CP) steps.}
\label{fig:PAM_Lossless}
\end{figure}

Let us combine the above two procedures, namely AP and CP, to design the lossless algorithm. To get a visual understanding of the algorithm, we showcase in Fig.~\ref{fig:PAM_Lossless} the process of generating the value of $P^3[D, E]$ without loss, for the original toy graph of Fig.~\ref{fig:PAM_rolling_example}. Specifically, two $3$-hop paths connect $D$ to $E$: $D \xrightarrow{3} A \xrightarrow{3} B \xrightarrow{5} E$ and $D \xrightarrow{3} C \xrightarrow{7} B \xrightarrow{5} E$, corresponding to relational chains $(r_1, r_1, r_2)$ and $(r_1, r_3, r_2)$ accordingly. 

At the top of the figure, we see the $1$-hop and $2$-hop lossless PAMs, alongside the $\phi_1, \phi_2$ and $\phi_3$ mappings. Below the dashed line, we present the procedure followed to generate the value of $P^3[D, E]$, starting by extracting the vectors $P^2[D,:]$ and $P[:, E]$, similarly to standard matrix multiplication. Then, we follow the chaining and aggregation procedure as described above, for each intermediate node $n_c$ that contains non-zero values (in this case, node $B$ is the only intermediate node). The resulting value is $57$, which can be decomposed to $57 = 3 \times 19$. These values can be further mapped to paths, as follows: $3 \xrightarrow{\phi{_3}^{-1}} (r_1, r_1, r_2)$ and $19 \xrightarrow{\phi{_3}^{-1}}  (r_1, r_3, r_2)$, which are indeed the exact relational chains that lead from $D$ to $E$ in $3$ hops.

In pseudocode the lossless algorithm to generate $P^{k+1}[i,j]$ is shown in Algorithm~\ref{algo:lossless_k_small}.

\begin{algorithm}[!htb]
   \caption{Lossless algorithm overview for $P^{k+1}[i,j]$}
    \hspace*{\algorithmicindent} \textbf{Input} $k$-hop vector $P^k[i, :]$, 1-hop vector $P[:, j]$, mapping functions $\phi_{1}, \phi_{k}, \phi_{k+1}$ \\
    \hspace*{\algorithmicindent} \textbf{Output}  Value of $P^{k+1}[i,j]$ 
\begin{algorithmic}[1]
\State $\tt{primes} = []$
\State $\tt{commonIndices} =$ \{$n_c: \forall n_c \in V, P^k[i,n_c]*P[n_c, j] \neq 0 $\} 

\For{$n_c \in \tt{commonIndices}$}
    \State $Paths^{k+1}[i,j] = \tt{Chaining Process(P^k[i,n_c]*P[n_c, j], \phi_{k}, \phi_{1})}$
    \State  $p_{n_c} = \tt{Aggregation Process(Paths^{k+1}[i,j], \phi_{k+1})}$
    \State $\tt{primes}.append(p_{n_c})$
\EndFor
\State $P^{k+1}[i,j]=\prod_{n_c}primes[n_c]$ if $\lvert \tt{commonIndices}\rvert > 0$ else 0
\end{algorithmic}
\label{algo:lossless_k_small}
\end{algorithm}

A more detailed algorithm and additional material regarding the $k$-hop lossless algorithm can be found in Appendix~\ref{sec:appendix_lossless}. In practice, deviating from standard matrix multiplication and performing the chaining and aggregation processes as proposed here, adds significant computational cost. With the current implementation, it is mainly useful for smaller graphs and a few $k$-hops.


To conclude, in this section we provided an overview of an algorithmic process for generating $k$-hop PAMs without loss of information. We achieved this by substituting the default operations of matrix multiplication with lossless path-chaining (instead of multiplication) and path-aggregation (instead of summation) processes. Our main interest in the proposed lossless approach is to build a framework that is interpretable by design. For instance, continuing from the illustrated example in Fig.~\ref{fig:PAM_Lossless}, where we have $P^3[D, E] = 57$, assume that the frequency of this value is found to be important for a downstream task, e.g. graph classification. Having generated the PAMs through the lossless algorithm allows us to backtrack the exact relational chains this value expresses, namely $(r_1, r_1, r_2)$ and $(r_1, r_3, r_2)$, gaining important insights related to the domain of the graph.

However, interpretability can be achieved for the default (lossy) PAMs as well. While we can't have the exact path decomposition for all the values in any given $P^k$, we can regenerate the exact paths given a subset of interest. For example, with the default scheme, we would have $P^3[D, E] = 150$. If the value $150$ is ``important'' to us, we can run a path extraction algorithm only for the pair of nodes $(D, E)$ that exhibit this value in $P^3$, which is computationally feasible\footnote{We can extract a single simple path between two given nodes in $\mathcal{O}(V+E)$.}. This would still allow us to find the same important relational chains as before, $(r_1, r_1, r_2)$ and $(r_1, r_3, r_2)$, gaining the same insights. Therefore, even when using the default lossy framework, we can generate interpretable insights with little computation overhead.

\subsection{Bag of Paths}\label{subsec:BoP}
With the PAM framework in place, we designed a simple feature extraction methodology that can be utilized for any downstream task on the graph, named \textit{Bag of Paths} (\textit{BoP}). Starting with a given graph, we calculate the PAMs up to the desired $k$-hop. Then, according to the downstream task, we generate feature vectors using these $k$-hop PAMs. This process involves simply extracting the nonzero values relevant to the task. These feature vectors compress and simplify the information available in the PAMs, retaining information from all the paths related to the entity of interest (i.e. node, edge, or subgraph) up to $k$-hops.

Formally, given the entity of interest $X$ (e.g. a node), we associate it with a subset of edges $E' \subseteq E$ from the graph that we consider related to the entity (e.g. the outgoing edges of a node). The Bag of Paths representation of this entity is defined as (for a given $k$):

\begin{equation}\label{eq:BoP_fundamental}
    BoP(X) = \mathop{\lvert \rvert }_{n=1}^{n=k} P^n[n_1, n_2], \forall (n_1, n_2) \in E' \textit{ if } P^n[n_1, n_2] \neq 0, 
\end{equation}
where the operator $\lvert \rvert $ concatenates the collected non-zero values. Essentially, for the edges of interest $E'$ related to the entity $X$, we aggregate the non-zero\footnote{In the rest of the paper, we may skip the check for zero values for brevity, but we still only keep non-zero elements.} values of the corresponding cells from all generated PAMs. 

\begin{figure}[htbp]
\centerline{\includegraphics[width=1\textwidth]{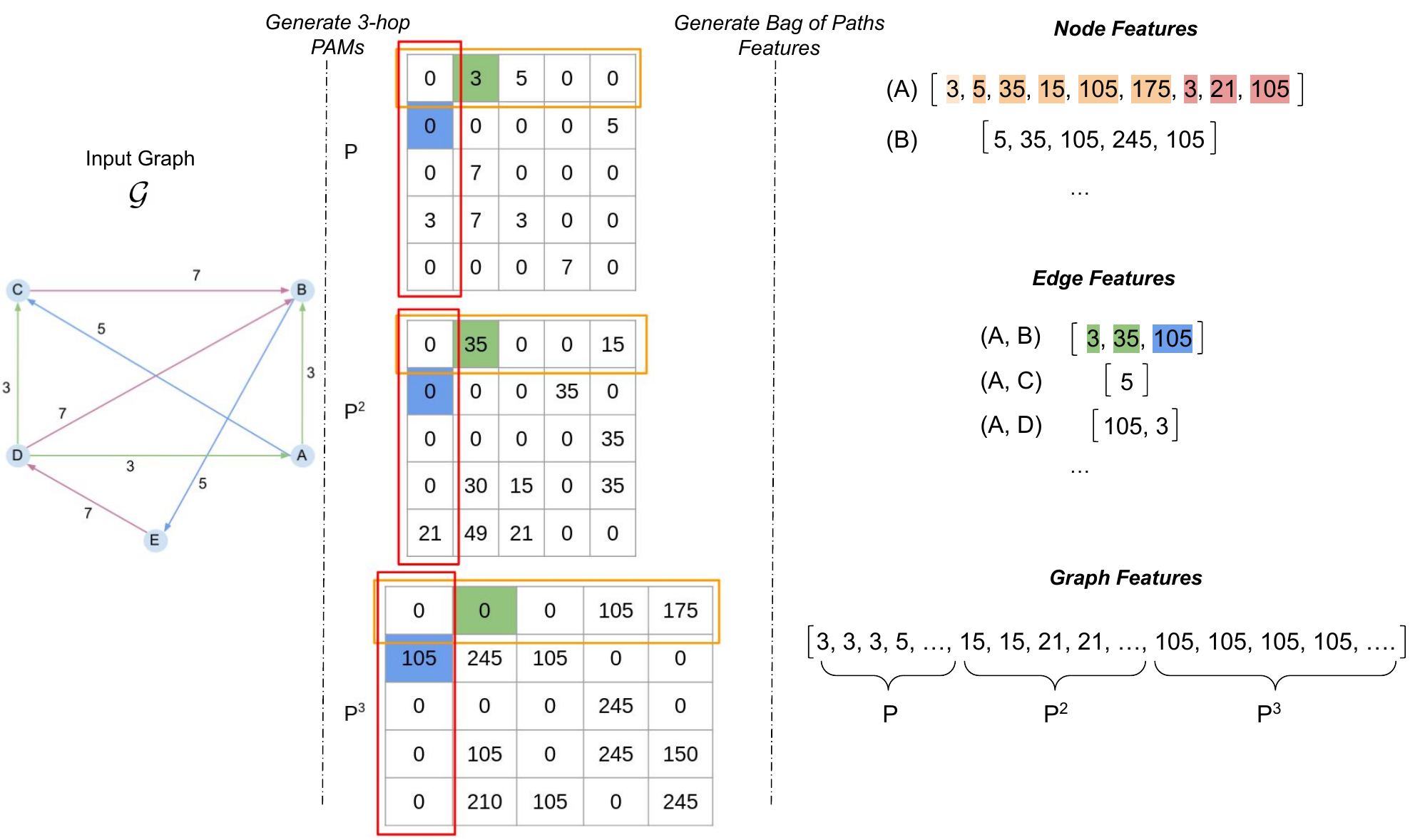}}
\caption{Illustration of the Bag of Paths extraction methodology across multiple scales, i.e. node-, edge- and graph-level tasks.}
\label{fig:PAM_BoP}
\end{figure}

The procedure is shown in Fig.~\ref{fig:PAM_BoP}, limited to $k=3$ for simplicity. Let us focus on a node classification task. Our goal is to generate a feature vector for each node. In this example, we focus on node A (i.e. the entity of interest), corresponding $P^n[0,:]$ and $P^n[:,0]$ columns and rows of the PAMs. Assume that the edges $E'$ that we believe are important to characterize the entity, are all the outgoing and incoming edges of the node. In this case Eq.~\eqref{eq:BoP_fundamental} would become:

\begin{equation}\label{eq:BoP_fundamental_node}
    BoP(A) = \mathop{\lvert \rvert }_{n=1}^{n=k}{P^n[0,:] \lvert \rvert  P^n[:,0]} = [3,5,35,15,105,175,3,21,105]
\end{equation}

Focusing on the corresponding orange rows and red columns of each PAM in Fig.~\ref{fig:PAM_BoP}, the non-zero element in each cell corresponds to a path that originates from (the orange ones) or terminates at (the red ones) node A. We aggregate these non-zero elements in a ``bag'' for each node, as shown at the upper right part of Fig.~\ref{fig:PAM_BoP} (the same procedure is shown for node B). The resulting bags can then be transformed into feature vectors with the frequencies of each path  (not depicted in the Figure). Essentially, each of these feature vectors would correspond to a histogram of paths for each node. These can then be used as input to a classifier for node classification.

Moving on to an edge-level task, we want to generate a feature vector for the pair \textit{(A, B)}, corresponding to $(0,1)$ in terms of matrix indices. For this given entity $X = (A,B)$,  assume that its related important edges $E'$ are all the paths connecting the pair. The resulting Bag of Paths representation for the pair would be: 

\begin{equation}\label{eq:BoP_fundamental_pair}
    BoP((A, B)) = \mathop{\lvert \rvert }_{n=1}^{n=k} P^n[0,1] \lvert \rvert P^n[1,0] = [3, 35, 105]
\end{equation}

Referring back to Fig.~\ref{fig:PAM_BoP}, we extract the paths that connect the head to the tail and vice versa, corresponding to the blue ($A\rightarrow B$) and green cells ($B\rightarrow A$) from all $k$-hop matrices. Aggregating these values, generates a feature vector with all the possible paths up to $k$-hops that connect $A$ and $B$  (middle right part of Fig~\ref{fig:PAM_BoP}). To increase expressivity, we can also augment the feature vector with the node features for the head (node A) and the tail (node B), which can be generated as shown above.

Finally, regarding graph-level tasks, we follow a similar procedure to generate one feature vector for the whole graph. In this case $X=\mathcal{G}$ and $E' = E$, meaning we keep all path values:

\begin{equation}\label{eq:BoP_fundamental_graph}
    BoP(\mathcal{G}) = \mathop{\lvert \rvert }_{n=1}^{n=k} P^n[n_1,n_2], \forall (n_1,n_2) \in E
\end{equation}

Thus, we simply extract all the non-zero elements from each PAM and aggregate them (bottom right part of Fig.~\ref{fig:PAM_BoP}), creating a feature vector that essentially captures the frequency of all paths up to $k$-hops in the graph.

In each of the above cases, we create feature vectors in the form of a Bag of Paths, containing counts of the paths expressed in different granularities: nodes, edges, and (sub)graphs. Inspired by natural language processing, we perform tf-idf weighting on collections of these BoP representations, akin to applying tf-idf to a collection of Bag-of-Words representations of documents. This process highlights informative paths, while filtering out very common path patterns in the collection. The resulting feature vectors can be used by a predictive model, e.g. a classifier,  related to the downstream task, e.g. classification, graph regression, etc.

An important characteristic of this general feature extraction methodology is that it creates interpretable features for the entities at hand, as each feature corresponds to a specific path and the value of the feature denotes the frequency of occurrence. This allows us to easily identify task-important paths, particularly when using a ``white-box'' classifier for the downstream task. 


\section{Applications}\label{sec:applications}

In the following subsections, we present multiple downstream tasks on graphs, utilizing PAMs and the BoP feature extraction methodology. All experiments were run on an Ubuntu Server with AMD Ryzen @ $3.9$GHz. We reserve $8$ threads and a maximum of $50$GB RAM for the experiments.

\subsection{Node Classification}\label{sec:application_nc}

The first task we address is node classification on heterogeneous graphs. To tackle this problem, we generate feature vectors for a given graph using PAMs and the BoP methodology, as described in Section~\ref{subsec:BoP}.

Specifically, for a given graph (dataset), we calculate the $k$-hop PAMs and generate the BoP representation for each node in the graph. Having these BoP representations for all the nodes, we apply tf-idf weighting over all paths in the collection (i.e., the values in the BoP representation), denoting the resulting vector as $F(x)$ for a given node $x$. Finally, we perform a $1$-hop feature aggregation around each node's neighborhood, formulating the final representation $H(x)$ as:

\begin{equation}\label{eq:nc}
H(x) = \alpha*F(x) + \frac{1}{|N(x)|}\sum_{n \in N(x)}{F(n)},
\end{equation}

\noindent where $\alpha$ is a parameter controlling the weight given to each node's representation and $N(x)$ are the $1$-hop neighbors of node $x$. The final feature vectors $H(x)$ serve as input to a CatBoost classifier~\citep{hancock2020catboost}. The values of $\alpha$ and $k$ (number of hops) are the only hyperparameters tuned for this model.

\begin{table}[h]
\centering
\caption{Characteristics of the datasets for node classification, including the number of classes for the target nodes, the number of relations in the graph, the total number of edges, the total number of nodes, and the number of train/test target nodes in each graph.}
\begin{tabularx}{1\linewidth}{lCCCCCC}
\toprule
Dataset & \# Classes & $\lvert R \rvert$ & $\lvert E \rvert$ &  $\lvert V \rvert$ & \# Train & \# Test \\
\midrule
AIFB & 4 & 45 & 29,043 & 8,285 & 140 & 36 \\
MUTAG & 2 & 23 & 74,227 & 23,644 & 272 & 68 \\
BGS & 2 & 103 & 916,199 & 333,845 & 117 & 29 \\
AM & 11 & 133 & 5,988,321 & 1,666,764 & 802 & 198 \\
\bottomrule
\end{tabularx}

\label{tab:nc_datasets}
\end{table}

We compare the performance of our model against state-of-the-art and other commonly-used models on four benchmark datasets~\citep{ristoski2016collection}: AIFB, MUTAG, BGS, and AM. As can be seen in Table~\ref{tab:nc_datasets}, the  datasets cover varying graph sizes, ranging from small (AIFB, with 8,285 nodes) to large (AM, with 1,666,764 nodes). We focus solely on the class labels of a small subset of nodes for each dataset (\textit{target nodes}), aiming to classify their classes correctly. The train and test splits are pre-determined and consistent with previous studies on these datasets. The competing models include the Weisfeiler-Lehman graph kernel (WL)~\citep{shervashidze2011weisfeiler}, a Relational Graph Convolutional Network (R-GCN)~\citep{https://doi.org/10.48550/arxiv.1703.06103}, a path-inducing tree-based method (WalkTree)~\citep{vandewiele2019inducing}, a neighborhood-based node embedding method (INK)~\citep{steenwinckel2022ink} and a recent GNN-based methodology, using cascaded convolutional layers (SCENE)~\citep{monninger2023scene}. Further details on the datasets and the experimental setup can be found in Appendix~\ref{sec:appendix_nc}.

\begin{table}[htbp]
\caption{Results on node classification in terms of accuracy (\%). The presented scores are averages over 5 runs and the best scores are in bold.}
\begin{tabularx}{1\linewidth}{l|CCCCCC}
\toprule
 Dataset & WL& R-GCN& WalkTree & SCENE & INK & BoP \\
\midrule
AIFB & 80.55 & \textbf{95.83} & 89.44 & \textbf{95.83} & 94.40 & 92.22 \\
MUTAG & 80.88 & 73.23 & 73.82 & 75.44 & 82.40 & \textbf{91.17 }\\
BGS & 86.20 & 86.20 & 86.90 & 92.41 & \textbf{93.10} & 90.34 \\
AM & 87.37 & 88.99 & 86.77 & 90.05 & 90.40 & \textbf{92.41} \\
\bottomrule
\end{tabularx}
\label{tab:nc_results}
\end{table}

The resulting scores are shown in Table~\ref{tab:nc_results}. The proposed methodology outperforms all competing models in two of the four datasets, MUTAG and AM. Specifically in MUTAG, which is also the hardest one based on all models' performance, we observe a relative increase of $10.64\%$ over the best performing model. The BoP model ranks third on the remaining two datasets, achieving a competitive performance overall. Moreover, it is important to note that the BoP model takes less than a minute on average across all datasets. Meanwhile, it only uses CPU resources making it a fast, resource-friendly approach that also achieves competitive results.

Our experiment shows that the proposed methodology is effective, even using simple path-based features as node descriptors and an out-of-the-box classifier. The objective was not to conduct extensive experimentation on node classification and propose a new state-of-the-art algorithms. Rather, we aimed to highlight the usefulness of the PAM framework, alongside the Bag of Paths representation, by employing a simple model that performs well.

\subsection{Relation Prediction}\label{sec:application_rp}

Moving on to an edge-level task, we focus on \textit{relation prediction}. This task involves predicting the most probable relation that should connect two nodes in a graph. Essentially, we aim to complete the triple $(h, ?, t)$ where \textit{h} is the head entity and \textit{t} is the tail entity, by connecting them with one of the available relations.

To create a representation for a given pair of nodes (h, t), we follow the same procedure as before, utilizing the Bag of Paths (BoP) feature vectors. Specifically, the feature vector for a pair (h, t) is the concatenation of the BoP representations for the pair, the head node, and the tail node. Formally, we express this as:

\begin{equation}\label{eq:bop_rp}
    {
        H(h, t) = [F(h, t) \lvert \rvert  F(t, h)\lvert \rvert  F(h) \lvert \rvert  F(t)]
    }
\end{equation}
where $F(h, t)$ denotes the BoP vector for the pair (as illustrated in the middle right part of Fig.~\ref{fig:PAM_BoP}), $F(h)$, $F(t)$ the BoP vectors for the head and tail (as illustrated at the top right of the same figure). As in the node classification tasks, we apply tf-idf weighting to the generated $H(h, t)$, using the entire collection of pair feature vectors. Finally, following the same approach as in~\citep{bougiatiotis_cn}, we use a simple k-Nearest-Neighbor (k-NN) model on these feature vectors to demonstrate the expressivity of the underlying feature space.

We evaluate this model following the experimental setup presented in~\citep{10.1145/3447548.3467247} and focus on the 3 most difficult KGs used in that study, namely: NELL995~\citep{xiong2017deeppath} which is a collection of triples extracted from the NELL system~\citep{carlson2010toward}, WN18RR~\citep{dettmers2018convolutional} based on WordNet, and DDB14 which is based on the Disease Database~\footnote{\url{http://www.diseasedatabase.com}}, a medical database containing biomedical entities and their relationships. The characteristics of the datasets used are summarized in Table~\ref{tab:datasets_rp}. 

\begin{table}[htbp]
\caption{Statistics of the datasets on relation prediction. $\mathbb E[d]$ and ${\rm Var}[d]$ correspond to the mean and variance of the node degree distribution.}
\begin{center}
\begin{tabularx}{1\linewidth}{lCCCCCCC}
\hline
\textbf{Dataset} & $\lvert V \rvert $ & $\lvert R \rvert$ & $\lvert E_{train}\rvert$ & $\lvert E_{val}\rvert$ & $\lvert E_{test}\rvert$ & $\mathbb E[d]$ & ${\rm Var}[d]$ \\
\hline
NELL995 & 63,917 & 198 & 137,465 & 5,000 & 5,000 & 4.3 & 750.6 \\
WN18RR & 40,493 & 11 & 86,835 & 3,034 & 3,134 & 4.2 & 64.3\\
DDB14 & 9,203 & 14 & 36,561 & 4,000 & 4,000 & 7.9 & 978.8\\
\hline
\end{tabularx}
\label{tab:datasets_rp}
\end{center}
\end{table}

We employ two variants of the BoP methodology: one using the default $k$-hop (PAMs) and another using the lossless algorithm for calculating PAMs, as proposed in Section~\ref{subsec:lossless_k_hop}. We report the results of competing well-known embedding models, as found in~\citep{10.1145/3447548.3467247}, alongside the \textit{PAM-knn}~\citep{bougiatiotis_cn}, which is also based on PAMs and employs a simpler version of feature extraction using PAMs.

We evaluate all models on the relation prediction task: for a given entity pair $(h, t)$ in the test set, we rank the ground-truth relation type $r$ against all other candidate relation types. We use \textit{MRR} (Mean Reciprocal Rank) and \textit{Hit@3} (hit ratio in the top-3 candidates) as evaluation metrics, consistent with the original work. More details on the competing models and their hyperparameters can be found in the original article, while in Appendix~\ref{sec:appendix_rp} further details on the BoP models are discussed.


\begin{table}[htbp]
\caption{Results for relation prediction on all datasets. The best results are highlighted in bold, while the second best is in italics. For DDB14 the number of parameters for each model is shown as well.}
\begin{tabularx}{1\linewidth}{l|CC|CC|CCC}
\toprule
\multirow{2}{*}{Model} &\multicolumn{2}{c|}{WN18RR} &\multicolumn{2}{c|}{NELL995} &\multicolumn{3}{c}{DDB14}  \\
 & MRR & H@3 & MRR & H@3 & MRR & H@3 & \# \nolinebreak Par. \\
\midrule
TransE & 0.784 & 0.870 & \textbf{0.841} & \textit{0.889} & \textit{0.966} &  \textit{0.980}  & 3.7M\\
CompleX & 0.840 & 0.880 & 0.703 & 0.765 & 0.953  & 0.968  & 7.4M\\
DistMult & 0.847 & 0.891 & 0.634  & 0.720 & 0.927 & 0.961  & 3.7M\\
RotatE & 0.799 & 0.823 & 0.729 & 0.756 & 0.953 & 0.964  & 7.4M\\
QuatE & 0.823 & 0.852 & 0.752 & 0.783 &  0.946 & 0.962 & 14.7M\\
\hline
PAM-knn & \textit{0.852} & \textit{0.957} & 0.740 & 0.843 & 0.915  & 0.961 & 0\\ 
\hline
BoP (default) & 0.816 & 0.946 & 0.775 & 0.848  & 0.949  & 0.978 & 0\\
BoP (lossless) &  \textbf{0.874} & \textbf{0.962} & \textit{0.816}& \textbf{0.891}  & \textbf{0.970}  & \textbf{0.988} & 0\\
\bottomrule
\end{tabularx}
\label{tab:rp_results}
\end{table}

The results of the experiments are presented in Table~\ref{tab:rp_results}. The lossless BoP variant outperforms all models in terms of Hit@3, while the default BoP performs better than most graph embedding methodologies on average across the datasets. This indicates that both representations are sufficiently expressive for the task, suggesting that the paths in the feature vectors hold important information for relation prediction. Additionally, the default variant with the BoP methodology outperforms the previous version of \textit{PAM-knn} in two of the three datasets, with a small performance drop in WN18RR.

\begin{figure}[htbp]
\centerline{\includegraphics[width=1\textwidth]{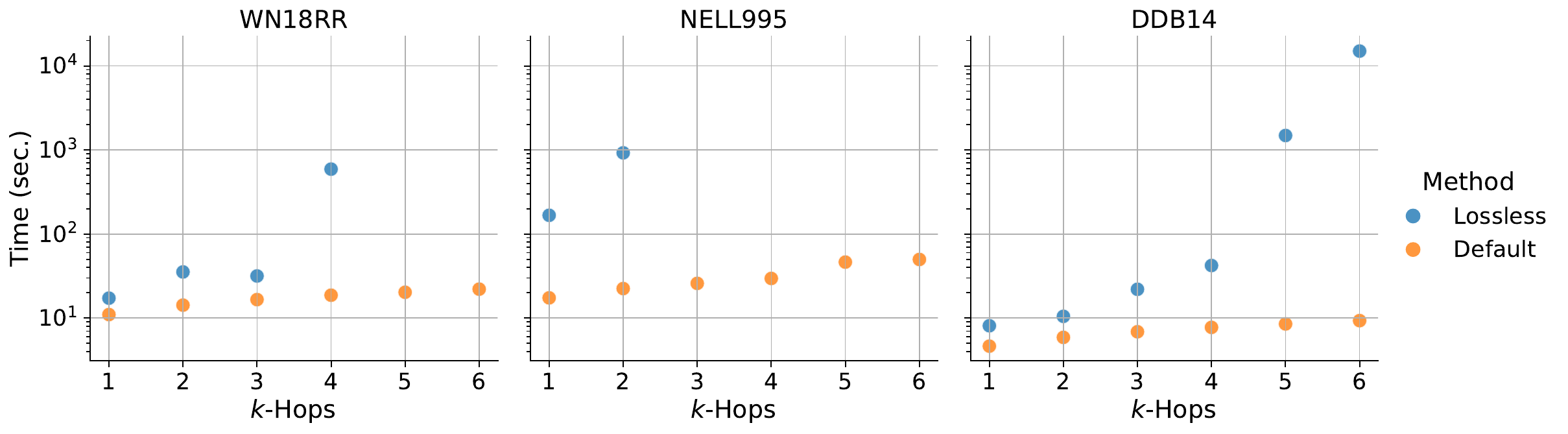}}
\caption{Time needed to run the two Bag of Paths variants across datasets, while varying the number of $k$-hop PAMs calculated. The y-axis is in log-scale and time is measured in seconds. Missing points indicate that the time needed exceeded 2.7 hours.}
\label{fig:rp_time}
\end{figure}

Once again it is important to note here that the PAM-based models, have no trainable parameters. This can be seen in the last column of the table, for DDB14  but the same is true for all datasets. This highlights the effectiveness of the representation, as the final predictive model is a simple nearest-neighbors model, relying solely on the expressivity of the space. Moreover, the entire process of training and inference takes a few minutes on all datasets using a CPU (for the lossless variant). At the same time, the embedding models were trained for hours using a GPU\footnote{According to the authors in~\citep{10.1145/3447548.3467247}, the models were trained for 1000 epochs each on a GPU and the best-performing model was selected, using the validation set.}. 

Finally, in Fig.~\ref{fig:rp_time} we illustrate the time needed for each BoP model when varying the number of $k$-hops across datasets. The default variant (orange) takes less than 10 seconds on the WN18RR and DDB14 datasets, even when $k$ goes up to 6, and around 100 seconds for NELL995. The time needed for all datasets exhibits a linear trend with a small slope on the log scale
For most practical applications, the actual time required is a matter of seconds or minutes, as demonstrated here. 

On the other hand, the lossless variant is more time-consuming, especially when $k > 2$. Specifically, for WN18RR ($k \geq 5$) and NELL995 ($k \geq 3$), the time required exceeds approximately $2.7$ hours (therefore no points are shown in the graphs for these cases). Moreover, the trendlines for the lossless data points appear to be either exponential or linear with a very steep slope on the log scale, making the current implementation of the model more suitable only for smaller graphs and low $k$ values.

Ultimately, the default variant of the BoP model offers a significant gain in terms of speed, compared to loss in performance for the relation prediction task.


\subsection{Graph Regression}\label{sec:application_gr}
Finally, for the graph-level task, we focus on graph regression. To generate a feature vector for a given graph, we follow the Bag of Paths feature extraction procedure illustrated at the bottom right of Fig.~\ref{fig:PAM_BoP}. Specifically, we first calculate the $k$-hop PAMs up to the desired $k$, then aggregate the non-zero elements found in the PAMs. These values represent all the different paths exhibited in the graph with a length of $k$ or less, and by keeping all occurrences of them, we also track their importance for each graph. Formally, this can be expressed as:

\begin{equation}\label{eq:bop_gr}
    {
        H(G) = \mathop{\lvert \rvert }_{n=1}^{n=k} P^n[i,j], \textit{ if } P^n[i,j] \neq 0 
    }
\end{equation}

Similarly to the procedure described in Sections 3.1 and 3.2, from a collection of graphs we first generate the corresponding $H(G)$ for each graph and then apply tf-idf weighting to generate the final graph feature vectors. These feature vectors are used as input to a CatBoost regressor. 

We evaluate the performance of the model on three commonly used benchmarks for graph regression: \textit{ZINC}, where we regress the constrained
solubility of ZINC molecules; \textit{AQSOL}, where we regress the aqueous solubility, both introduced in \citep{dwivedi2023benchmarking}; and the \textit{Peptides-struct} dataset from~\citep{dwivedi2022long}  which requires capturing long-range dependencies between nodes. For Peptides-struct, we regress on 11 targets for each molecule corresponding mainly to 3D properties. Details regarding the characteristics of the datasets can be seen in the first columns of Table~\ref{tab:gr_datasets}.

In all three datasets, the nodes represent heavy atoms, and the edges represent different types of bonds connecting them. To capture the different molecular properties expressed in these bonds, in the $1$-hop PAM we combine the information about the head and tail node types, as well as the bond connecting them. Specifically, each unique triple \textit{type(head)}-\textit{bond}-\textit{type(tail)} is assigned a unique prime number. For example, we have \textit{C}-\textit{Single}-\textit{C} $\rightarrow 3$, \textit{C}-\textit{Double}-\textit{C} $\rightarrow 5$, etc., with \textit{C} denoting a carbon atom and \textit{Single}/\textit{Double} denoting the bond type. Thus, the values in the $1$-hop PAM express these unique connections between atoms, while the values in $k$-hop PAMs represent $k$-hop bond chains.

We compare our methodology with several commonly used graph neural networks (GNNs): the Graph Attention Network (GAT)\citep{velivckovic2017graph}, the Graph Isomorphism Network (GIN)\citep{xu2018powerful},
and a more complex model from the family of Graph Transformers (GT)~\citep{dwivedi2020generalization}. For GAT and GIN, we report scores using models with approximately 100K parameters; for GT, the number of parameters is approximately 500K. In comparison, our proposed BoP method uses 10K features as input to the CatBoost regressor.


The scores for the competing models are sourced from ~\citep{dwivedi2022long, gabrielsson2023rewiring}, and we follow the same experimental setup. The evaluation metric used is Mean Absolute Error (MAE), where lower is better, and the average score over 4 runs is reported. For the \textit{Peptides-struct} dataset which has 11 regression variables, the score corresponds to the average MAE across all variables. 

\begin{table}[h]
\centering
\caption{Dataset statistics including the number of targets, training, validation, testing samples, and performance metrics of various models. The scores are average MAE$(\downarrow)$ over 4 runs with different seeds. With \textbf{bold} the best-performing scores.}
\label{tab:gr_datasets}
\begin{tabularx}{1.01\linewidth}{l|P{0.8cm}CCC|CCCP{0.9cm}}
\toprule
Dataset & Targets & Train & Valid. & Test  & GAT & GIN & GT & BoP \\
\midrule
ZINC & 1 & 1,0000 & 1,000 & 1,000 & 0.475 & 0.387 & 0.598 & \textbf{0.297} \\
AQSOL & 1 & 7,800 & 996 & 996 & 1.441 & 1.894 & \textbf{1.110} & 1.320  \\
Peptides-struct & 11 & 10,700 & 2,300 & 2,300 & 0.353 & 0.364 & 0.253 &\textbf{0.250}\\
\bottomrule
\end{tabularx}
\end{table}

Table~\ref{tab:gr_datasets} presents the performance of the models, alongside some characteristics of the datasets. At first glance, we can see that the proposed simple model clearly outperforms the GNN-based baselines GAT and GIN on all three datasets. The BoP model also outperforms the Graph Transformer in two out of three datasets. Specifically, for the \textit{Peptides-struct} dataset, our performance is within $3\%$ of the current SotA model~\citep{he2023generalization}. 

These results highlight the expressivity of the BoP feature extraction methodology and thus the usefulness of the PAM framework in graph-level tasks. Once again, these results were achieved in a matter of minutes using only CPU computations, while the competing models require several epochs on GPUs\footnote{The benchmarks in \citep{dwivedi2023benchmarking} were run on a server with 4 Nvidia 1080 Ti GPUs (11GB each).}.

Finally, due to the interpretable nature of the values in PAMs and the simplicity of the BoP methodology, we can interrogate the important paths for each task. In Table~\ref{tab:gr_expl} we see the top-$5$ correlated path features to the constrained solubility target variable for the \textit{ZINC} dataset.  The second row showcases the value $77$ (as found in the PAMs) with a correlation to the target variable of approximately $0.61$. This path value can be factorized to $77 = 7 \times11$ which denotes a $2-$hop chain. The first hop has a value of $7$, corresponding to the triple \textit{(C - Single - C)}, as seen in the first row of the same table. The second hop is $11$ corresponding to the triple \textit{(C - Double - C)}. Concatenating these hops creates the full chain, (\textit{C - Single - C} - \textit{Double - C}).

\begin{table}[h]
\centering
\caption{Top-$5$ correlated paths to the target variable for \textit{ZINC}. The first column corresponds to the value of the path as found in the PAMs, the second column corresponds to its prime decomposition, the third column maps the primes to the original path and the last column contains the Pearson correlation of the path with the target variable.}
\begin{tabularx}{1\linewidth}{ccP{5cm}c}
\toprule
Value & Factorization & Path Derivation & Importance \\
\midrule
7 & 7 & \textit{(C - Single - C)} & 0.624122 \\
77 & 7 $\times$ 11 & (\textit{C - Single - C} - \textit{Double - C}) & 0.612344 \\
43 & 43 & \textit{(C - Single - NH1+)} & -0.483504 \\
47 & 47 & \textit{(NH1+ - Single - C)} & -0.422128 \\
1046 & 523 + 523 = 2*523 &  2*\textit{(Br - Single - C H1)} & 0.364646 \\
\bottomrule
\end{tabularx}
\label{tab:gr_expl}
\end{table}

The first two rows that indicate highly correlated paths essentially capture the fact that carbon-carbon bonds decrease solubility by imparting hydrophobicity to organic molecules. These bonds lack significant polarity and do not readily interact with water. Additionally, longer carbon chains and molecules with numerous carbon-carbon bonds exhibit heightened hydrophobic character, further reducing their solubility in aqueous environments, by limiting their interaction with water molecules~\citep{cooper2020clue}.

Therefore, using PAMs, we can easily generate insights by focusing on task-important paths. This, combined with competitive performance when measured against other commonly used models and the minimal time and computational resources required, makes the proposed approach a valuable alternative for graph-level tasks.

\section{Related Work}\label{sec:related_work}


Understanding the structural properties of multi-relational graphs is essential for tackling various downstream tasks such as link prediction, multi-hop reasoning, graph question answering, and rule learning, among others. Path-based and structure-aware methodologies provide a promising direction to this end, by leveraging the rich connectivity patterns that uncover latent relationships and improve performance. This section reviews the key frameworks and methods that utilize path-based information for multi-relational graphs, focusing on approaches used for reasoning and learning.

\subsection{Reasoning}

\textbf{Path-based Techniques}: The Path Ranking Algorithm (PRA)~\citep{lao2011random} is one of the earliest methods that leverages paths as symbolic features for link prediction. While effective, PRA is computationally intensive, especially for large graphs. Path-RNN~\citep{neelakantan2015compositional} and PathCon~\citep{10.1145/3447548.3467247} improve PRA by learning representations of paths with recurrent neural networks (RNNs), reducing the computational burden. However, these methods typically operate on a limited set of paths, while also restricting the length to avoid exponential growth. Chains of Reasoning~\citep{das2016chains} and All-Paths~\citep{toutanova2016compositional} adopt dynamic programming techniques to efficiently aggregate all paths between entities, though All-Paths has limited model capacity. Recent GNN-based models such as RED-GNN~\citep{zhang2022knowledge}, PathNN~\citep{michel2023path}, and A*Net~\citep{zhu2024net} further explore the use of paths and GNNs. Specifically, A*Net searches for crucial paths instead of utilizing all potential paths for reasoning, using an A*-inspired heuristic. Despite great improvements in performance, these methods are still computationally demanding\footnote{A*Net was trained using 4 Tesla A100 (40GB) GPUs.}.

\textbf{Reinforcement Learning Methods}: Methods like DeepPath~\citep{xiong2017deeppath} and MINERVA~\citep{das2017go} use reinforcement learning to navigate and collect meaningful paths in a knowledge graph. These approaches address the infeasibility of exhaustive path enumeration by guiding the exploration process. However, they suffer from sparse rewards and often limit the number of hops, which can restrict their effectiveness. Moreover, the need to balance exploration and exploitation in large action spaces further complicates their application.

\subsection{Learning}

\textbf{Association Rule Mining}: Traditional path-body rule learning approaches like AMIE~\citep{galarraga2013amie} face scalability challenges due to the vast rule space. More recent methods such as AnyBURL~\citep{meilicke2024anytime} and RARL~\citep{pirro2020relatedness} address this by sampling rule instances, thus accelerating the learning process. AnyBURL for example uses policy-guided sampling and approximates rule confidence using only a subset of the knowledge graph. These improvements are crucial to make the methods usable in large KGs, nevertheless, they sacrifice exploration for speed.

\textbf{Reinforcement Learning Methods}: Methods like RNNLogic~\citep{qu2020rnnlogic} and RLogic~\citep{cheng2022rlogic} use reinforcement learning to derive logical rules. RNNLogic struggles to scale with the number of relations and entities, while RLogic relies on a small set of sampled closed paths for training, which can limit its generalizability. R5~\citep{lu2022r5} aims to improve rule learning efficiency, but shares similar limitations with the above approaches, particularly in handling the large and complex search space.

\textbf{Tensor Reasoning}: Methods based on the tensor reasoning framework TensorLog~\citep{cohen2016tensorlog}, represent logical inference problems as sequences of differentiable tensor multiplications. NeuralLP~\citep{yang2017differentiable} employs an attention mechanism to guide the search space of possible tensor multiplications, while DRUM~\citep{sadeghian2019drum} uses an RNN to share information when learning rules for different relations. Although these methods transform discrete optimization into differentiable problems, they still face scalability issues when applied to large graphs.

\subsection{Discussion}

As presented in Section~\ref{sec:related_work}, many of the current path-based methods suffer from scalability issues, the need for heuristics, or guided exploration to find important paths for the task at hand. The PAM framework addresses many of these limitations by enabling efficient computation of the $k$-hop paths between all nodes through the $k$-hop PAMs, without the need for sampling or heuristics. This inherent efficiency makes PAMs particularly suitable for real-world applications, potentially enhancing both inference and learning tasks in complex networks in the following ways:\newline

\textbf{PAM as a tool}: A straightforward way is to utilize the proposed Bag of Paths methodology for the task at hand, in similar ways as demonstrated in Section~\ref{sec:applications}. Alternatively, one could use the PAMs directly to address the downstream task. For example, one way to utilize PAMs for rule mining would be to generate rules/patterns $M_i$ that infer relationships of the form:

    \begin{equation}
        {
         M_i : r_1(X, A_1) \wedge r_2(A_1, A_2) \wedge \ldots \wedge r_k(A_{k-1}, Y) \rightarrow r_{H}(X,Y) 
        }
    \end{equation}
    
\noindent where the above rule states that if there is a $k$-hop chain of relations $r_1, r_2, ..., r_k$ from $X$ to $Y$, we can infer that they should be connected with the relation $r_H$ directly. Using PAMs, this chain of relations would simply be expressed as: $$P^k[X,Y]=\prod_{i=1}^{i=k}p(r_i),$$ where $p(r_i)$ is the prime allocated to relation $r_i$. With this understanding, we can generate all such $k$-length possible rules $M_i$ by extracting all pairs of nodes $(X,Y)$ that have non-zero values in both $P^1$ and $P^k$. Formally this is expressed as follows:

    \begin{equation}
         M_i :  {Path}^k(X,Y) \rightarrow r_{H}(X,Y) \overset{Eq.\eqref{eq:paths_mapped_product}}{\iff} P^k[X,Y]\rightarrow P[X,Y]
    \end{equation}

\noindent Thus, through look-ups in PAMs we can easily generate rules of the form $P^k[X,Y]\rightarrow P[X,Y]$ and calculate their exact confidence in a matter of minutes.\newline

\textbf{PAM as a component}: Another approach would be to take advantage of the time efficiency of PAMs to enhance existing methodologies. Specifically, the framework could substitute a component that significantly increases complexity or is crucial for the target task. For example, in AnyBURL~\citep{meilicke2024anytime} the confidence estimation of rules is done using only very few ground instances of each rule. Using PAMs, we could easily check through look-ups whether a rule is satisfied over all its ground bodies. Another example the A*Net~\citep{zhu2024net}, where instead of the priority function being used, one could guide the system to important paths per relation, utilizing rules directly generated from PAMs.

A different way to utilize PAMs is to instill structural awareness in existing methods either through the BoP features or directly through PAMs, at very low cost. A case of the former would be to augment existing node features with the structurally-rich BoP node features of Section~\ref{sec:application_nc}, before using an off-the-self GNN model for node classification. An example of the latter would be to utilize PAMs to find important paths for node classification (e.g. by correlating BoP features with node labels) as a pre-processing step in a graph-rewiring-based methodology as in \citep{guo2023homophily}.

To conclude, path-based methods often face limitations such as the computational expense of the path search space, sampling biases, and instability due to heuristics. The PAM framework could mitigate these issues by providing direct access to all multi-hop paths through simple matrix operations, therefore eliminating the need for heuristic/guided path exploration or random sampling.

\section{Conclusions}\label{sec:conclusions}
In this work, we focused on the Prime Adjacency Matrix representation for dealing with multi-relational networks. It is a compact representation that incorporates information from multiple relations in a single-adjacency matrix. This, in turn, leads to efficient ways of generating higher-order adjacency matrices, that encapsulate the rich structural information of the graph. We further augmented the original framework with a lossless algorithm for generating the higher-order matrices and showcased its applicability in graph applications. 

We also designed a fast feature extraction methodology, named Bag of Paths, that enables versatile and fast feature extraction for node-, edge-, and graph-level tasks. Moreover, the generated feature vectors are interpretable by design, aiding in the identification of task-important paths found in the graphs. The effectiveness of these contributions has been demonstrated through competitive performance in various applications, all achieved with minimal computational resources.

The PAM framework  can be enhanced in various ways. Extending the framework to handle weighted and dynamic multi-relational networks presents an exciting research direction. Further optimization techniques, such as parallel processing and advanced matrix factorization methods, could significantly reduce computation time, making the framework even more practical for large-scale applications and  enabling the use of the lossless algorithm at scale. Additionally, validating PAM across a broader range of domains and applications will help understand its generalizability and robustness. Finally, it is interesting to see which ideas that are useful in single-relational analysis can be applied to PAMs as well, such as spectral analysis~\citep{spielman2012spectral}, topological graph theory~\citep{gross2001topological} and others.


\bibliography{main}
\clearpage
\appendix

\section{GraphBLAS Implementation}\label{sec:appendix_graphblas}

To showcase the impact of using GraphBLAS~\citep{erik_welch_2024_10631255} in PAM for matrix multiplications, we benchmarked the time required for some commonly used knowledge graphs. Specifically, we experimented on WN18RR~\citep{dettmers2018convolutional}, YAGO3-10-DR~\citep{akrami2020realistic}, FB15k-237~\citep{toutanova2015representing}, NELL995~\citep{xiong2017deeppath}, CoDex-L~\citep{safavi2020codex}, and HetioNet~\citep{himmelstein2017systematic}. The first four are well-known datasets for link prediction in knowledge graphs, each with distinct structural characteristics. CoDEx-L is the largest variant of the CoDEx dataset from WikiData. Finally, HetioNet is significantly larger, in terms of the number of edges, than all the other datasets used here. It is a biomedical network with multiple node and relation types, and many hub nodes, making the generation of powers of its adjacency matrix quite challenging.

The basic characteristics of the six datasets are shown in Table~\ref{tab:scalability}, including the number of nodes, the number of unique types of relations, and the total number of edges (in the training set). We also present the time (in seconds) required to calculate each PAM of order $k=5$ using the previous SciPy-based implementation from~\citep{bougiatiotis_cn} and the current implementation with GraphBLAS. The last column shows the resulting speedup in performance, expressed as a ratio. For each dataset, $5$ runs were made and the reported times are the average for each method.

\begin{table}[htbp]
\caption{Main characteristics of the knowledge graphs and the time needed to calculate the corresponding $P^5$ PAMs using the two different implementations, alongside the achieved $\times$Speedup with GraphBLAS, as a ratio.}
\begin{center}
\begin{tabularx}{0.98\linewidth}{lCCC|CCC}
\toprule
Dataset & $\lvert V\rvert$ & $\lvert R\rvert$ & $\lvert E\rvert$  &\multicolumn{3}{c}{$P^5$ (sec.)}\\
 & &  & & SciPy & GraphBLAS & Ratio\\
\midrule
WN18RR & 40,493 & 11 & 86,835 & 0.16 & 0.06 & $\times$2.5 \\
NELL995 & 57,016 & 199 & 118,304 & 6.82 & 0.55 & $\times$12.58\\
CoDEx-L  & 77,951 & 69 & 551,193 & 5.51 & 0.39 & $\times$14.15\\
FB15k-237 & 14,541 & 237  & 272,115 & 12.78 & 0.34 & $\times$37.87  \\
YAGO3-10-DR & 122,837 & 36 & 732,556 & 14.01 & 0.87 & $\times$16.06  \\
HetioNet & 45,158 & 24 & 2,250,197 & 72.07 & 4.23 & $\times$17.05 \\
\bottomrule
\end{tabularx}
\label{tab:scalability}
\end{center}
\end{table}

\begin{figure}[htbp]
\centerline{\includegraphics[width=1\textwidth]{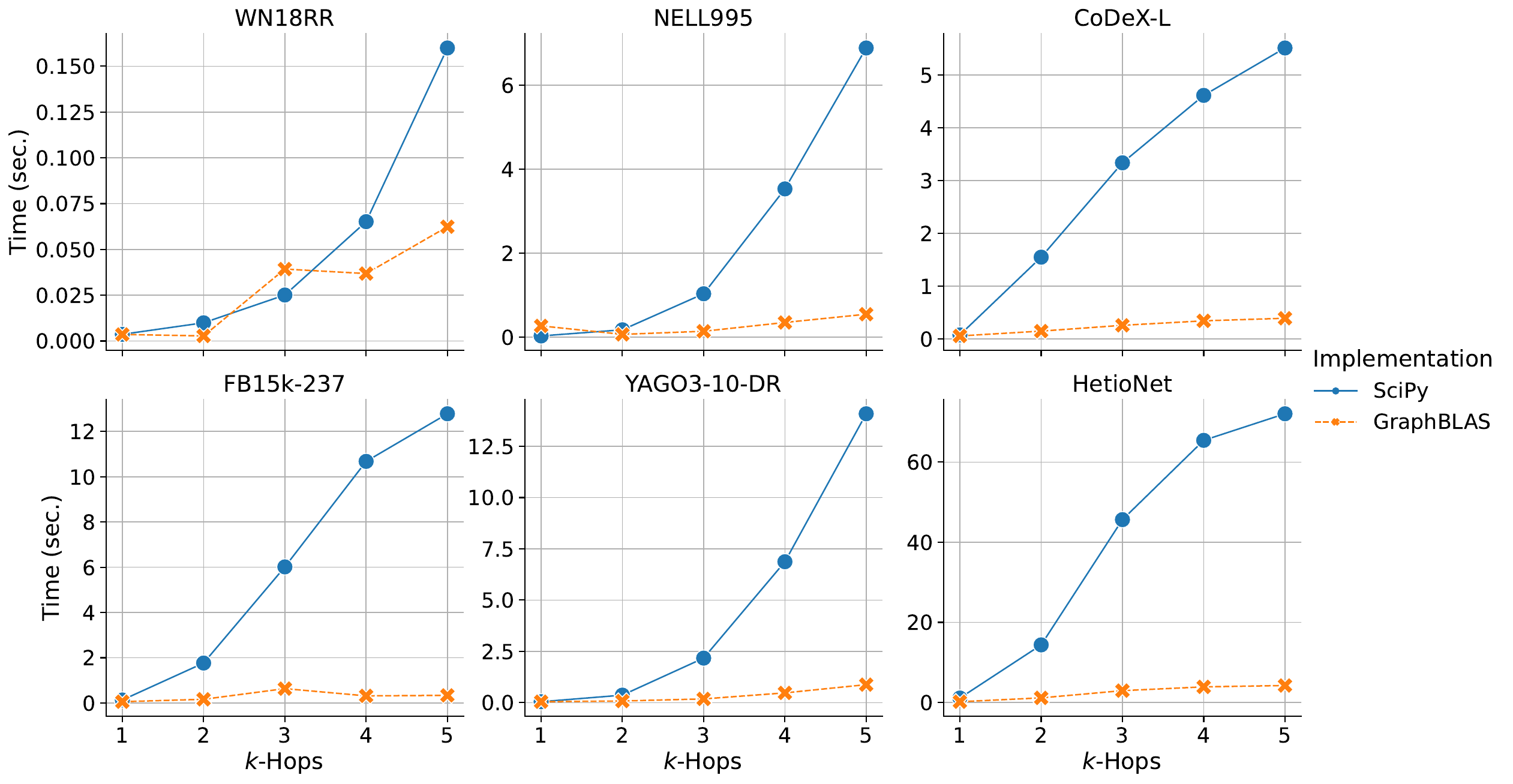}}
\caption{Time needed to generate the $k$-hop PAMs across different knowledge graphs, using the GraphBLAS implementation versus the SciPy-based from~\citep{bougiatiotis_cn}.}
\label{fig:scalability}
\end{figure}


Overall, the performance with the GraphBLAS implementation is significantly improved, with calculations more than 10 times faster for medium- and large-scale knowledge graphs. Moreover, as seen in Fig.~\ref{fig:scalability}, the speedup becomes more pronounced, as the order $k$ of the PAM increases. In conclusion the substantial enhancement in performance demonstrates the technical advantage of the updated version with the GraphBLAS backend and highlights its potential impact on large-scale applications in various domains.

\section{Lossless k-hop}\label{sec:appendix_lossless}
First we define the Fundamental Theorem of Arithmetic (FTA):

\begin{theorem}[Fundamental Theorem of Arithmetic - FTA]\label{theorem:FTA}
Every positive integer $n > 1$ can be represented in exactly one way as a product of prime powers:
\begin{equation}
    n = p_1^{n1}p_2^{n2}...p_k^{nk} = \mathop{\prod}_{i=1}^k p_i,
\end{equation}
where $p_1 < p_2 < ... < p_k$ are primes and $n_i$ positive integers.
\end{theorem}

Following the workflow presented in Section~\ref{subsec:lossless_k_hop}, we will prove the losslessness of the algorithm through induction.

\noindent\underline{\textit{Base case: k=1}}\newline

Recall that for the original PAM matrix (with $k=1$) we have:
\begin{equation}\label{eq:PAM_base}
        P[i,j]=
        \begin{cases}
             \displaystyle \prod_{r:A[i,j,r]=1} p_r & if\text{ $\exists r : A[i,j,r]=1 $} \vspace{0.3em}\\ 
             \hspace{2em} $0$ &if\text{ $\forall r :$ $A[i,j,r]=0$},
        \end{cases}
\end{equation}

where each relation $r$ in the graph is mapped to the prime $p_r$, through the mapping function $\phi_1$\footnote{As noted, this mapping function sequentially maps relations to primes 2, 3, 5, etc. Its inverse function is denoted as $prime2path^1$.}.

\noindent Thus, for all the non-zero values $p[i,j]$: 
\begin{equation}\label{eq:pij}
    p[i,j] = \prod_{\texttt{r connects i to j}} p_r, 
\end{equation}

\noindent which is the product of all the 1-hop paths (relations) that connect $i$ to $j$. Based on Eq.~\eqref{eq:pij} and Eq.~\eqref{eq:paths_mapped_product} for $k=1$, we infer
that these values are the same. Therefore, we can uniquely re-create the collection of 1-hop paths (relations) that connect $i$ to $j$ given $p[i,j]$ and the $prime2path^1$ dictionary (i.e. the inverse of the $\phi_1$ mapping function) $\square$.

\vspace{3pt}
\noindent\underline{\textit{Induction step}}\newline

Let us assume now that $P^k$  has been created, using the proposed algorithm. $P^k$ maintains all the information about paths of $k$-hops. We will prove that the same holds for the case of $P^{k+1}$. If it holds for the case of $P^k$, this means that we can also have access to $prime2path^k$, which maps each prime in a distinct path of $k$-hops.

In order to construct, $P^{k+1}[i,j]$, we need all the $(k+1)$-hop paths that connect $i$ to $j$:

\begin{equation}\label{eq:pij_k1_appendix}
    P^{k+1}[i,j] = \{path_1, path_2 ...\} =\{\overbrace{(r_1, r_2, ...)}^{k+1}, \overbrace{(r_1, r_2, ...)}^{k+1}, ...\},
\end{equation}

Each of these $(k+1)$-hop paths can be broken into 2 parts, the first of which denotes the  $k$-hop path starting from node $i$ and reaching an intermediate node $n_c$, and the second one being the final hop from $n_c$ to the tail node $j$, i.e.:

\begin{equation}\label{eq:path_segm_appendix}
    path_i = \overbrace{(r_1, r_2, ...)}^{k+1} = \overbrace{(r_1,r_2,...,r_k)}^{part-1} \lvert\rvert \overbrace{(r_{k+1})}^{part-2}, 
\end{equation}

where $\lvert\rvert$ denotes the concatenation of the two paths.

\noindent We can collect all $(k+1)$-hop paths that connect $i$ to $j$ by finding all $k$-hop paths starting from $i$ and leading to any $n_c$ in $\mathcal{G}$, with the constraint that there is also a 1-hop connection between $n_c$ and $j$:

\begin{equation}\label{eq:path_k+1_def_appendix}
\small
Paths^{k+1}[i,j] = \{n_c \in V, P^k[i,n_c]*P[n_c, j] \neq 0: Paths^k[i,n_c]\lvert \rvert Paths^1[n_c,j]\},
\end{equation}

Using $P^k$, we can recreate the collection of paths $Paths^k[i,n_c]$, by factorizing $P^k[i,n_c]$ and using $prime2path^k$ to map these prime factors to $k$-hop paths. This can be done $\forall n_c \in V$. Similarly, as we proved in the base case for $k=1$, we can recreate the collection of paths $Paths^1[n_c,j]$ using $P[n_c,j]$ and $prime2path^1$.

Therefore, we have access to both $Paths^k[i,n_c]$ and $Paths^1[n_c,j]$ $\forall i,j,n_c \in V$, and we can calculate every possible $path \in Paths^{k+1}[i,j]$ according to Eq.~\eqref{eq:path_k+1_def_appendix}. For a given $[i,j]$ pair, this is done through:
\begin{enumerate}
    \item Iterating over their common nodes $n_c$.
    \item Concatenating all possible combinations  between the $k$-hop paths in $Paths^k[i,n_c]$ and the 1-hop paths in $Paths^1[n_c,j]$ to generate all \mbox{$(k+1)$-hop} paths that connect $i$ and $j$ through $n_c$.
    \item Mapping each of these distinct $(k+1)$-hop paths to a corresponding distinct prime.
    \item Calculating the product (i.e., $P^{k+1}[i,j]$) of all these primes that correspond to the paths in the collection (i.e., $Paths^{k+1}[i,j]$).
\end{enumerate}

By repeating this procedure for all $i,j$ pairs, we construct the $P^{k+1}$ matrix, where all non-zero values $P^{k+1}[i,j]$ are products of primes that can be mapped back to the exact collection of $(k+1)$-hop paths that connect $i$ and $j$ $\square$. \newline

A detailed algorithm that describes the previous workflow for the entire $P^{k+1}$ PAM matrix is shown in Algorithm~(\ref{algo:lossless_k}). It expects as input the order $k=1$ PAM $P$ (along with the corresponding mapping of primes to 1-hop paths/relations in $prime2path^1$) and the $P^k$ PAM (along with the corresponding mapping of primes to $k$-hop paths in $prime2path^k$)\footnote{These $prime2path^k$ dictionaries are essentially the inverse of the prime mapping functions $\phi_1, \phi_k$, which map the paths to the primes.}. With these as input, it outputs the $P^{k+1}$ PAM (alongside the corresponding mapping of primes to $(k+1)$-hop paths in $prime2path^{(k+1)}$).

\begin{algorithm}[!htb]
   \caption{Lossless Generation procedure for $P^{k+1}$}
    \hspace*{\algorithmicindent} \textbf{Input} PAM 1-hop matrix $P$, PAM k-hop matrix $P^k$, dictionary mapping the primes in $P$ to the original 1-hop paths (relations) $\tt{prime2path^1}$, dictionary mapping the primes in $P^k$ to paths with k-hops $\tt{prime2path^k}$, mapping function from paths of (k+1) hops to primes $\phi_{k+1}$ \\
    \hspace*{\algorithmicindent} \textbf{Output}  PAM (k+1)-hop matrix $P^{k+1}$, dictionary mapping the primes in $P^{k+1}$ to paths with (k+1)-hops $\tt{prime2path^{k+1}}$
\begin{algorithmic}[1]
   
   \For{$i \in range(0, N)$}
    \For{$j \in range(0, N)$}
        \State $\tt{commonIndices} =$ the non-zero overlapping indexes of $P^k[i,:]$ and $P[:,j]$
        \State $\tt{product} = 1$
        \For{$n_c \in \tt{commonIndices}$}
            \State $\tt{primeFactors^k} = factorize(P^k[i,n_c])$
            \State $\tt{primeFactors^1} = factorize(P[n_c,j])$
            \For{$\tt{primeFactorPath} \in \tt{primeFactors^k}$}
                \State $\tt{path^k} = \tt{prime2path^k[primeFactorPath]}$
                \For{$\tt{primeFactorHop} \in \tt{primeFactors^1}$}
                    \State $\tt{relation} = prime2path^1[primeFactorHop]$
                    \State $\tt{path^{k+1}} = $ $\tt{path^k}$  $\lvert \rvert $ $\tt{relation}$
                    \State $\tt{currentPrime} = \phi_{k+1}(path^{k+1})$
                    \State $\tt{product} = \tt{product} * \tt{currentPrime}$
                    \State $\tt{prime2path^{k+1}[currentPrime]} = \tt{path^{k+1}} $
                 \EndFor
            \EndFor
        \EndFor
            \State $P^{k+1}[i,j] = \tt{product}$ if $\tt{product}$ $\neq 1$ else $0$
    \EndFor
   \EndFor
\end{algorithmic}
\label{algo:lossless_k}
\end{algorithm}

\vspace{3pt}

\section{Applications}

\subsection{Node Classification}\label{sec:appendix_nc}

\noindent \textbf{Bag of Paths}  The Bag of Paths representation for  node $x$ with matrix index $i$ is:
\begin{equation}\label{eq:BoP_node_appendix}
    F(x) = F_{out}(x) \lvert \rvert F_{inc}(x)
\end{equation}

\noindent The representation $F_{out}(x)$ corresponds to the outgoing paths expressed by the node (i.e. highlighted as orange values in Fig.~\ref{fig:PAM_BoP} for node A): 

\begin{equation}
    F_{out}(x) = \mathop{\lvert \rvert }_{n=1}^{n=k} P^n[i,:], \textit{ if } P^n[i,:] \neq 0, 
\end{equation}

\noindent while $F_{inc}$ the incoming paths of the node ( highlighted as red values in Fig.~\ref{fig:PAM_BoP} for node A):

\begin{equation}
    F_{inc}(x) = \mathop{\lvert \rvert }_{n=1}^{n=k} P^n[:,i], \textit{ if } P^n[:,i] \neq 0, 
\end{equation}

\noindent $F(x)$ is a feature vector where each feature corresponds to a specific path, and its value is the number of times this path either starts at or ends to the node.

Having generated the final feature vectors for each node:

$$ H(x) = \alpha*F(x) + \frac{1}{|N(x)|}\sum_{n \in N(x)}{F(n)} $$,

\noindent we perform a tf-idf weighting procedure on the feature values. During this process, we limit the paths that we are focusing on by the following rules:
\begin{itemize}
    \item A path exhibited by less than $2$ nodes is skipped, as it is deemed too rare.
    \item A path exhibited by more than $99$\% of the nodes is skipped, as it is deemed too common.
    \item We keep only the top 10,000  paths, in terms of frequency, for efficiency.
\end{itemize}
Due to this final rule, the final node feature vectors forwarded to CatBoost, will have (at most) a size of 10,000.\newline

\noindent \textbf{Hyper-parameters} Regarding the hyper-parameters $k$ (i.e. the number of k-hop PAMs to use) and $\alpha$ (i.e. the self-weighting parameter) in $H(x)$, the ranges checked and the selected values per dataset are shown in Table~\ref{tab:nc_hyper}:

\begin{table}[h]
\centering
\caption{Ranges and selected values per dataset for the hyper-parameters $k$ and $\alpha$.}
\label{tab:nc_hyper}
\begin{tabularx}{0.99\linewidth}{l|CCCCC}
\toprule
Parameter & Range &  AIFB & MUTAG & BGS & AM \\
\midrule
$k$ & \{1,2, ..., 10\} & 4 & 3 & 2 & 2  \\
$\alpha$ & \{1,2,5,10\} & 2 & 2 & 5 & 5 \\
\bottomrule
\end{tabularx}
\end{table}

For CatBoost, we use the default hyper-parameters, but we use also reserve $10$\% of each training set as a validation split, to be used for early stopping of the model, with a patience of $20$ epochs. The same validation split is used to select the values of $k$ and $\alpha$ for each dataset as shown in Table~\ref{tab:nc_hyper}.

\subsection{Relation Prediction}\label{sec:appendix_rp}

\noindent \textbf{Bag of Paths}  The BoP features generated for the pair $(h,t)$ is :
\begin{equation}
    H(h, t) = [F(h, t) \lvert \rvert  F(t, h) \lvert \rvert  F(h) \lvert \rvert  F(t)],
\end{equation}

\noindent where $F(h)$, $F(t)$ are the feature vectors for the head and tail nodes, as generated in the node classification scenario (i.e. as seen in Eq.~\eqref{eq:BoP_node_appendix} before the neighborhood aggregation). 
$F(h, t)$ and $F(t, h)$ are the feature vectors of each (directed) pair as:

\begin{equation}
    F(h, t) =  \mathop{\lvert \rvert }_{n=1}^{n=k} P^n[h,t], \textit{ if } P^n[h,t] \neq 0,
\end{equation}

\noindent Essentially, the values in $F(h, t)$ aggregate all the possible paths of length up to $k$ that connect node $h$ to $t$, as they correspond to all the non-zero entries in the $k$-hop PAMs.\newline

\noindent \textbf{Evaluation setup}  The first evaluation measure that we use is \textit{Mean Reciprocal Rank} ($MRR$) as :

\begin{equation}
    MRR =  \frac{1}{\lvert Q \rvert}\sum_{i=1}^{i=Q}\frac{1}{rank(r_q)},
\end{equation}

\noindent where $\lvert Q \rvert$ is the number of test triples $(h_q, r_q, t_q)$ and $rank(r_q)$ is the rank of the query relation $r_q$, when sorting all possible relations that complete $(h_q, ?, t_q)$.

In the same context, \textit{Hits at 3} ($H@3$) corresponds to:

\begin{equation}
    H@3 =  \frac{1}{\lvert Q \rvert}\sum_{i=1}^{i=Q} \mathbbm{1}[rank(r_q) \leq 3],
\end{equation}
where $\mathbbm{1}[rank(r_q) \leq 3]$ is $1$ if $rank(r_q) \leq 3$, else $0$.
\newline

\noindent \textbf{Hyper-parameters} Regarding the hyper-parameters $k$ (i.e. the number of k-hop PAMs to use) and $N$ (i.e. the number of neighbors used in the k-NN model) the ranges checked and the selected values per dataset are shown in Table~\ref{tab:rp_hyper}:

\begin{table}[h]
\centering
\caption{Ranges and selected values per dataset for the hyper-parameters $k$ and $N$, for each of the two Bag of Paths variants.}
\label{tab:rp_hyper}
\begin{tabularx}{0.99\linewidth}{lC|CCCC}
\toprule
Variant & Parameter & Range &  WN18RR & NELL95 & DDB14 \\
\midrule
Default & $k$ & \{1,2, ..., 6\} & 5 & 4 & 5   \\

Lossless & $k$ & \{1,2, ..., 6\} & 2 & 2 & 3   \\

Default & $N$ & \{5,10,20,50,100\} & 100 & 20 & 20  \\
Lossless & $N$ & \{5,10,20,50,100\} & 100 & 100 & 20  \\
\bottomrule
\end{tabularx}
\end{table}

The best-performing parameters were selected using the dedicated validation split in each dataset. Interestingly, the lossless variant requires fewer hops (lower $k$) to reach its optimal performance across all datasets compared to the default (lossy) variant.

\subsection{Graph Regression}\label{sec:appendix_gr}

\noindent \textbf{Hyper-parameters} Regarding the hyper-parameter $k$, which is the only one used in this task, the ranges checked and the selected values per dataset are shown in Table~\ref{tab:gr_hyper}:

\begin{table}[h]
\centering
\caption{Ranges and selected values per dataset for the hyper-parameter $k$ for the Bag of Paths model.}
\label{tab:gr_hyper}
\begin{tabularx}{0.99\linewidth}{l|CCCCC}
\toprule
Parameter & Range &  ZINC & AQSOL & Peptides-struct \\
\midrule
$k$ & \{1,2, ..., 10\} & 5 & 5 & 6 \\
\bottomrule
\end{tabularx}
\end{table}

The best-performing parameters were selected using the dedicated validation split in each dataset.



\end{document}